\documentclass[letterpaper, 10 pt, conference]{ieeeconf}  

\IEEEoverridecommandlockouts                              

\usepackage{amsmath,amsfonts}
\usepackage{algorithmic}
\usepackage{colortbl}
\usepackage{placeins}
\usepackage{xcolor}
\usepackage{algorithm}
\usepackage{array}
\usepackage[caption=false,font=normalsize,labelfont=sf,textfont=sf]{subfig}
\usepackage{textcomp}
\usepackage{orcidlink}
\usepackage{stfloats}
\usepackage{url}
\usepackage{verbatim}
\usepackage{orcidlink}
\usepackage{graphicx}
\usepackage{makecell}
\usepackage{xcolor}
\usepackage{booktabs,multirow,pifont}
\usepackage{cite}
\usepackage{float}
\usepackage{siunitx}
\usepackage{xcolor}
\usepackage{colortbl}
\usepackage{balance}
\usepackage{xurl}

\title{\LARGE \bf 
DiffusionAnything: End-to-End In-context Diffusion Learning for Unified Navigation and Pre-Grasp Motion
}
\author{%
    \begin{minipage}{\linewidth}
    \centering
    Iana~Zhura,
    Yara~Mahmoud,
    Jeffrin~Sam,
    Hung~Khang~Nguyen,
    Didar~Seyidov,
    Miguel~Altamirano~Cabrera,
    Dzmitry~Tsetserukou
    \thanks{The authors are with the Intelligent Space Robotics Laboratory, Center for Digital Engineering, Skolkovo Institute of Science and Technology.
    {\tt \{yana.zhura, yara.mahmoud, jeffrin.sam, khang.nguyen, didar.seyidov, m.altamirano, d.tsetserukou\}@skoltech.ru}}
    \end{minipage}
}
\begin{document}

\maketitle
\thispagestyle{empty}
\pagestyle{empty}


\begin{abstract}
Efficiently predicting motion plans directly from vision remains a fundamental challenge in robotics, where planning typically requires explicit goal specification and task-specific design. Recent vision-language-action (VLA) models infer actions directly from visual input but demand massive computational resources, extensive training data, and fail zero-shot in novel scenes. We present a unified image-space diffusion policy handling both meter-scale navigation and centimeter-scale manipulation via multi-scale feature modulation, with only 5\,min of self-supervised data per task. Three key innovations drive the framework: (1) Multi-scale FiLM conditioning on task mode, depth scale, and spatial attention enables task-appropriate behavior in a single model; (2) trajectory-aligned depth prediction focuses metric 3D reasoning along generated waypoints; (3) self-supervised attention from AnyTraverse enables goal-directed inference without vision-language models and depth sensors. Operating purely from RGB input (2.0\,GB memory, 10\,Hz), the model achieves robust zero-shot generalization to novel scenes while remaining suitable for onboard deployment.
\end{abstract}

\section{Introduction}

The deployment of versatile robotic systems, such as mobile manipulators and humanoids, requires a seamless integration of mobility and dexterity. Traditionally, navigation and manipulation have been treated as largely independent domains \cite{kroemer2021review}. Navigation systems focus on meter-scale collision avoidance and global routing \cite{Zhang2022survey}, while manipulation requires centimeter-scale precision, dynamic force regulation, and complex kinematic planning \cite{chi2023diffusion}. This dichotomy often results in modular, cascaded architectures that suffer from compounding errors, high latency, and rigid hand-offs between operational modes \cite{stone2023ok, rana2023sayplan}.
\begin{figure}[t]
\centering
\includegraphics[width=\columnwidth]{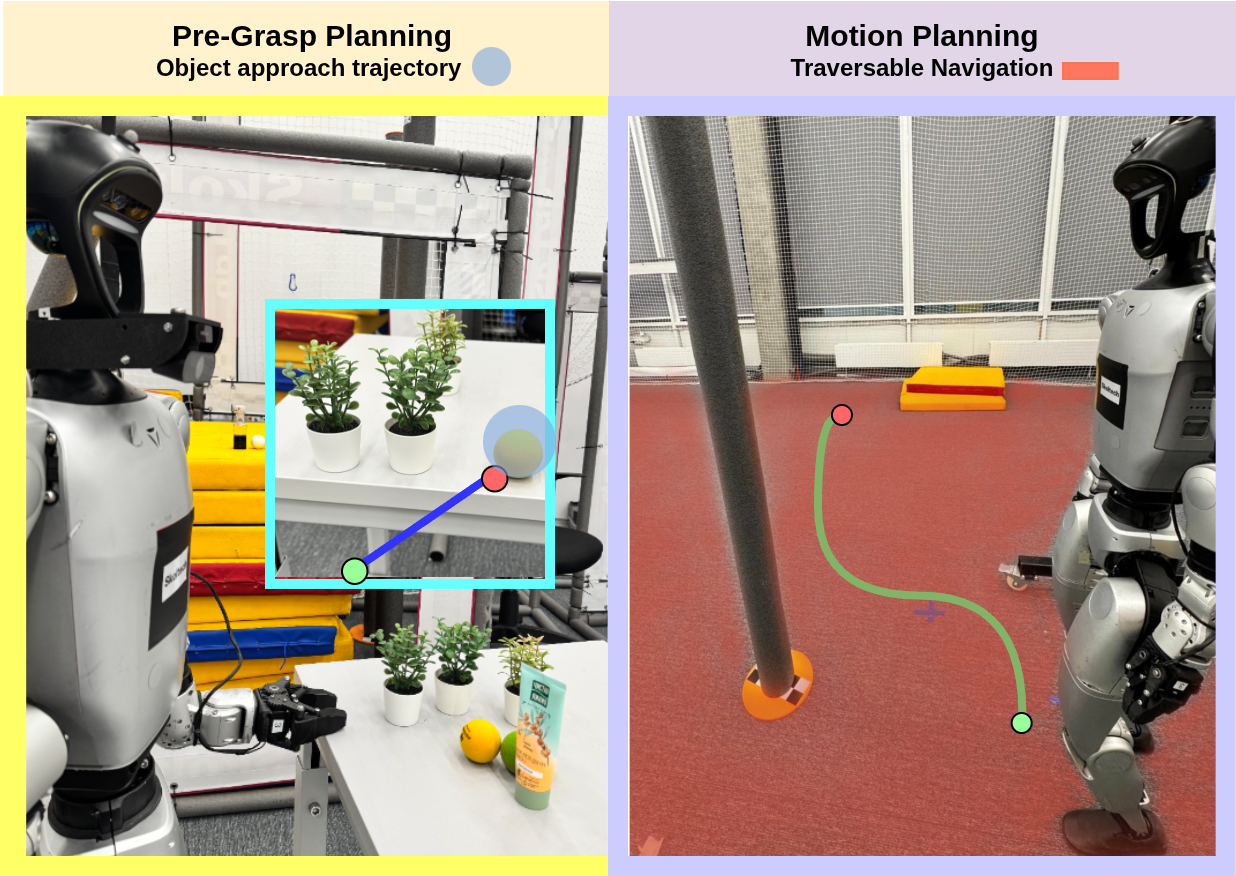}
\caption{A single unified diffusion policy handles both centimeter-scale pre-grasping planning (left: object approach with $\pm$2\,cm precision, cyan inset shows object attention) and meter-scale motion planning (right: hallway navigation with $\pm$10\,cm obstacle avoidance, orange overlay shows traversable floor). Knowledge transfer between tasks is achieved through context-aware conditioning with shared model weights.}
\label{fig:teaser}
\end{figure}
Recently, the paradigm has shifted toward end-to-end learning with the advent of Vision-Language-Action (VLA) foundation models \cite{brohan2022rt1}. Architectures such as RT-2 \cite{brohan2023rt2}, OpenVLA \cite{kim2024openvla}, and Project GR00T~\cite{gr00tn1_2025} attempt to unify diverse robotic tasks by casting them as sequence modeling problems. While these models demonstrate impressive zero-shot generalization and semantic reasoning, they come with prohibitive computational costs \cite{Firoozi2025fundationm}. They demand massive, heterogeneous datasets for training and rely on billions of parameters, resulting in high inference latency that is often incompatible with the high-frequency control loops required for real-time physical interaction and pre-grasping maneuvers. Furthermore, their reliance on language conditioning can be cumbersome for tasks that are inherently spatial or geometric \cite{ze2024gnfactor}.

To bridge the gap between meter-scale mobility and precise, object-centric pre-grasping planning without the overhead of massive VLA models, we propose DiffusionAnything. Instead of relying on heavy transformer-based semantic reasoning, our approach leverages a lightweight, context-aware diffusion policy.




In this work, we propose a unified diffusion-based framework that addresses the limitations mentioned above. Our key contributions are:

\begin{itemize}
    \item \textbf{Context-Aware Cross-Task Diffusion Architecture}: A unified diffusion policy that handles both collision-free navigation at meter-scale and object-centric pre-grasping motion planning at centimeter-scale through multi-scale FiLM conditioning on task mode, depth scale, and spatial attention. This context learning framework enables a single model to exhibit task-appropriate behaviors while operating purely from RGB at inference. The model performs obstacle avoidance and goal-directed trajectory prediction for navigation, while generating precise object approach trajectories for pre-grasping motion planning.
    
    \item \textbf{Trajectory-Aligned Depth Reasoning}: Efficient 3D reasoning through depth prediction exclusively along generated trajectory waypoints rather than dense depth maps. Combined with depth-scale conditioning, this enables metric planning for both collision-free navigation (meter-scale obstacle distances) and centimeter-precise pre-grasping trajectories within a single framework, reducing computational overhead while maintaining task-relevant geometric accuracy.
    
    \item \textbf{Adaptive Attention Prediction for Goal Learning}: Lightweight attention predictor learned from AnyTraverse supervision generates task-appropriate spatial focus (object-centric for manipulation, goal-directed for navigation, floor regions for exploration). Supports dual operational modes: autonomous goal selection for trained categories and zero-shot goal specification for novel objects via visual feature matching with reference images. Eliminates heavy vision-language models and depth sensors while enabling flexible goal-directed behavior without retraining.
\end{itemize}

\section{Related Work}
\subsection{Learning-Based Trajectory Planning for Manipulation}

Traditional approaches to pre-grasp trajectory planning have predominantly relied on geometric planning methods and optimization-based techniques. Works such as Pre-Grasp Manipulation~\cite{avigal2022pre} formulate the problem in configuration space using sampling-based planners, which require explicit obstacle representations and struggle with complex, cluttered environments. Similarly, motion planning frameworks like MoveIt~\cite{moveit} combine collision checking with inverse kinematics solvers, but these methods are brittle when faced with partial observability and dynamic scenes.

Recent learning-based approaches have shown promise in overcoming these limitations. Diffusion Policy~\cite{chi2023diffusion} demonstrates that diffusion models can generate smooth, multimodal action distributions for visuomotor control, achieving state-of-the-art performance on manipulation benchmarks. However, this work operates primarily in joint-space control and requires task-specific training for each manipulation primitive. 3D Diffuser Actor~\cite{ke20243d} extends diffusion to 3D keypoint-based manipulation with language conditioning, but relies on explicit 3D scene reconstruction and keypoint detection, which can be noisy and computationally expensive in real-world scenarios.

SE(3)-Diffusion~\cite{urain2023se3} addresses trajectory generation on Riemannian manifolds for task-space planning, providing theoretically grounded methods for handling SE(3) poses. While elegant, this approach requires precise state estimation and does not naturally handle high-dimensional sensory inputs like RGB images. M$\pi$Nets~\cite{hansen2023mpinets} combines diffusion with motion primitives for bimanual manipulation, but the reliance on hand-crafted primitives limits generalization to novel pre-grasping scenarios.

A key limitation across these manipulation-focused diffusion works is their \textbf{task-specific nature}: models trained for grasping cannot be directly applied to navigation or other robotics tasks, necessitating separate models and training pipelines for each capability. Additionally, most methods operate in explicit state spaces (joint angles, Cartesian poses, or 3D keypoints) rather than directly in image space, requiring intermediate perception modules that introduce cascading errors.

\subsection{Goal-Conditioned Visual Navigation}

In the navigation domain, goal-conditioned policies have achieved remarkable success. Navidiffusor~\cite{zeng2025navidiffusor} and NoMaD~\cite{sridhar2023nomad} demonstrate diffusion-based navigation policies conditioned on goal images, showing robust performance across diverse environments. GNM (General Navigation Model)~\cite{shah2022gnm} learns distance-based navigation from heterogeneous datasets, enabling zero-shot deployment in novel environments. However, these navigation policies are fundamentally designed for mobile base control and cannot be directly applied to manipulation tasks requiring precise end-effector control.

The navigation community has largely operated independently from manipulation research, resulting in \textbf{separate model architectures, training procedures, and deployment pipelines}. This separation is inefficient from both computational and data perspectives, as both tasks fundamentally involve goal-directed trajectory generation in visual environments. Recent works like SayPlan~\cite{rana2023sayplan} and OK-Robot~\cite{stone2023ok} attempt to combine navigation and manipulation but rely on modular pipelines with separate components, lacking the end-to-end unification we propose.

\section{Technical Approach}
\subsection{Network Architecture}

\begin{figure*}[t]
\centering
\includegraphics[width=0.65\textwidth]{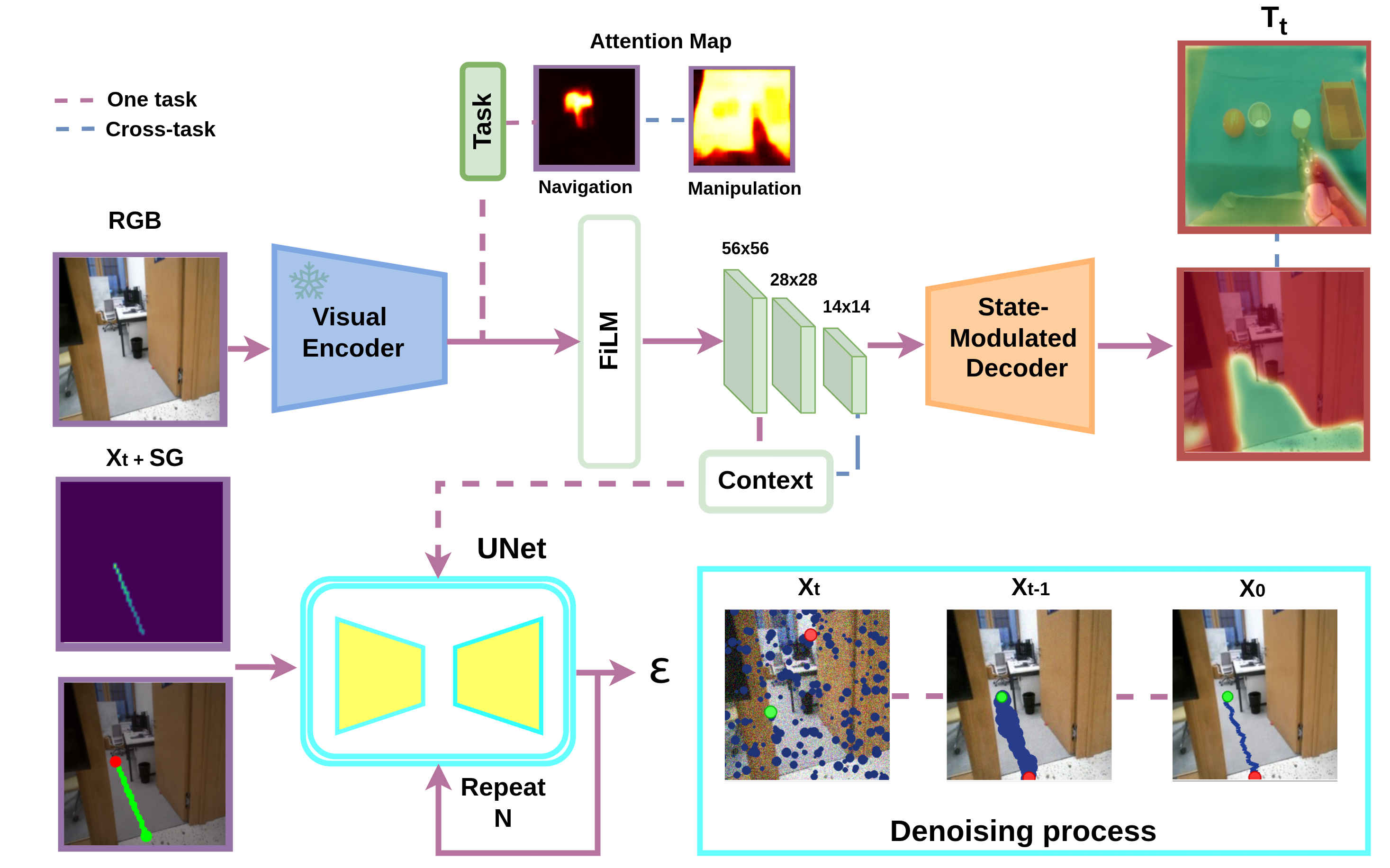}
\caption{Context-aware cross-task diffusion policy architecture. \textbf{Top (Supervision):} RGB input is processed by a frozen visual encoder, modulated via FiLM conditioning using task-specific context (task mode, depth scale, spatial attention). The state-modulated decoder outputs traversability $\hat{\mathbf{T}}_t$ for supervision. Attention maps differ by task: navigation uses floor traversability (left heatmap), while manipulation focuses on target objects (right heatmap). Purple dashed lines indicate single-task paths; blue dashed lines show cross-task conditioning shared between operational modes. \textbf{Bottom (Diffusion):} The UNet performs iterative denoising conditioned on context $\mathbf{c}$ and noisy trajectory $\mathbf{x}_t + \mathbf{S}_g$, progressively refining predictions over $N$ steps (right: $\mathbf{x}_t \rightarrow \mathbf{x}_{t-1} \rightarrow \mathbf{x}_0$).}
\label{fig:architecture}
\end{figure*}
\subsubsection{Diffusion policy}

Building on recent advances in traversability-guided diffusion for robotic planning~\cite{zhura2025swarmdiffusion}, we inherit key design principles: utilizing learned traversability representations for diffusion conditioning and image-space trajectory formulations. We extend this foundation to tackle cross-task generalization across navigation and pre-grasping motion planning, along with zero-shot adaptation to novel scene configurations. The core diffusion policy predicts a denoised trajectory $\mathbf{x}_0$ from noisy input $\mathbf{x}_t$ conditioned on RGB observations and context:
\begin{equation}
\epsilon_\theta(\mathbf{x}_t, t, \mathbf{I}, \mathbf{c}) \rightarrow \mathbf{x}_0,
\end{equation}
where $t$ is the diffusion timestep, $\mathbf{I}$ is the RGB input, and $\mathbf{c}$ is the context vector encoding task mode, depth scale, and spatial attention. 

Unlike the original embodiment-agnostic framework designed for heterogeneous robot navigation, our architecture introduces multi-scale FiLM conditioning to handle fundamentally different behavioral modes within a single model. The UNet backbone processes visual features through an encoder-decoder structure with skip connections, while context-aware modulation at multiple scales enables the model to adapt feature representations for coarse global navigation versus fine-grained pre-grasp motion planning tasks.

During training, the model learns to denoise trajectories through the standard diffusion objective:
\begin{equation}
\mathcal{L}_{\text{diff}} = \mathbb{E}_{t, \mathbf{x}_0, \epsilon} \left[ \|\epsilon - \epsilon_\theta(\mathbf{x}_t, t, \mathbf{I}, \mathbf{c})\|^2 \right],
\end{equation}
where $\epsilon \sim \mathcal{N}(0, \mathbf{I})$ is the noise added at timestep $t$. At inference, we perform iterative denoising starting from pure noise $\mathbf{x}_T \sim \mathcal{N}(0, \mathbf{I})$ using DDPM sampling.
\subsubsection{Context learning via multi-scale feature modulation}

To distinguish between navigation and pre-grasping within a unified architecture, we introduce context learning through Feature-wise Linear Modulation (FiLM) applied at multiple UNet encoder scales.

\textbf{Context representation.} Our context comprises three signals: (1) Task mode $m \in \{0, 1\}$ (navigation vs pre-grasping), (2) Depth scale $s \in \{\text{meter}, \text{cm}\}$ for metric interpretation, and (3) Spatial attention $\mathbf{A} \in \mathbb{R}^{64 \times 64}$ highlighting task-relevant regions. For navigation, $\mathbf{A}$ encodes traversable floor regions; for pre-grasping motion planing, $\mathbf{A}$ encodes object attention from AnyTraverse. These are combined into a conditioning vector:
\begin{equation}
\mathbf{c} = [\mathbf{e}_m \;||\; \mathbf{e}_s \;||\; \text{Enc}_{\text{attn}}(\mathbf{A})],
\end{equation}
where $\mathbf{e}_m$, $\mathbf{e}_s$ are learned embeddings and $\text{Enc}_{\text{attn}}$ is a lightweight convolutional encoder.

\textbf{Multi-Scale FiLM.} We apply FiLM at four encoder scales ($64 \times 64$, $32 \times 32$, $16 \times 16$, $8 \times 8$):
\begin{equation}
\mathbf{f}_\ell' = \gamma_\ell(\mathbf{c}, \mathbf{A}_\ell) \odot \mathbf{f}_\ell + \beta_\ell(\mathbf{c}, \mathbf{A}_\ell),
\end{equation}
where $\mathbf{A}_\ell$ is the attention map downsampled to scale $\ell$. To incorporate spatial guidance, we use spatially-varying modulation:
\begin{equation}
\gamma_\ell(\mathbf{c}, \mathbf{A}_\ell) = \gamma_\ell^{\text{global}}(\mathbf{c}) \cdot (1 + \mathbf{M}_\ell(\mathbf{A}_\ell))
\end{equation}
\begin{equation}
\beta_\ell(\mathbf{c}, \mathbf{A}_\ell) = \beta_\ell^{\text{global}}(\mathbf{c}) \cdot \mathbf{M}_\ell(\mathbf{A}_\ell),
\end{equation}
where $\mathbf{M}_\ell$ generates spatial masks from attention. This enables task-specific feature emphasis: pre-grasping task amplifies object regions at fine scales, while navigation enhances free-space features at coarse scales.

\textbf{Learned scale weighting.} To enable adaptive multi-scale emphasis, we introduce learnable task-scale weights $\mathbf{W} \in \mathbb{R}^{K \times L}$ where $K$ is the number of tasks and $L$ is the number of scales. For task $k$, encoder features at scale $\ell$ are weighted as:
\begin{equation}
\mathbf{f}_\ell^{(k)} = w_{k,\ell} \cdot \mathbf{f}_\ell.
\end{equation}
We initialize $\mathbf{W}$ to reflect task-scale affinities: pre-grasping tasks receive higher weights at fine scales ($w_{k,\text{fine}} = 1.0$, $w_{k,\text{coarse}} = 0.5$), while navigation tasks prioritize coarse scales ($w_{k,\text{fine}} = 0.5$, $w_{k,\text{coarse}} = 1.0$). These weights are optimized end-to-end during training, allowing the model to learn optimal scale emphasis for each operational mode.

\subsubsection{Goal learning via attention guidance}

Goals and start positions are predicted as diffusion output channels, enabling end-to-end spatial reasoning. A ViT encoder processes RGB observations to extract semantic features, which predict both traversability maps and task-specific attention maps (object-centric for manipulation, goal regions for navigation, floor regions for exploration).

\begin{figure*}[t]
    \centering
    \includegraphics[width=0.8\textwidth]{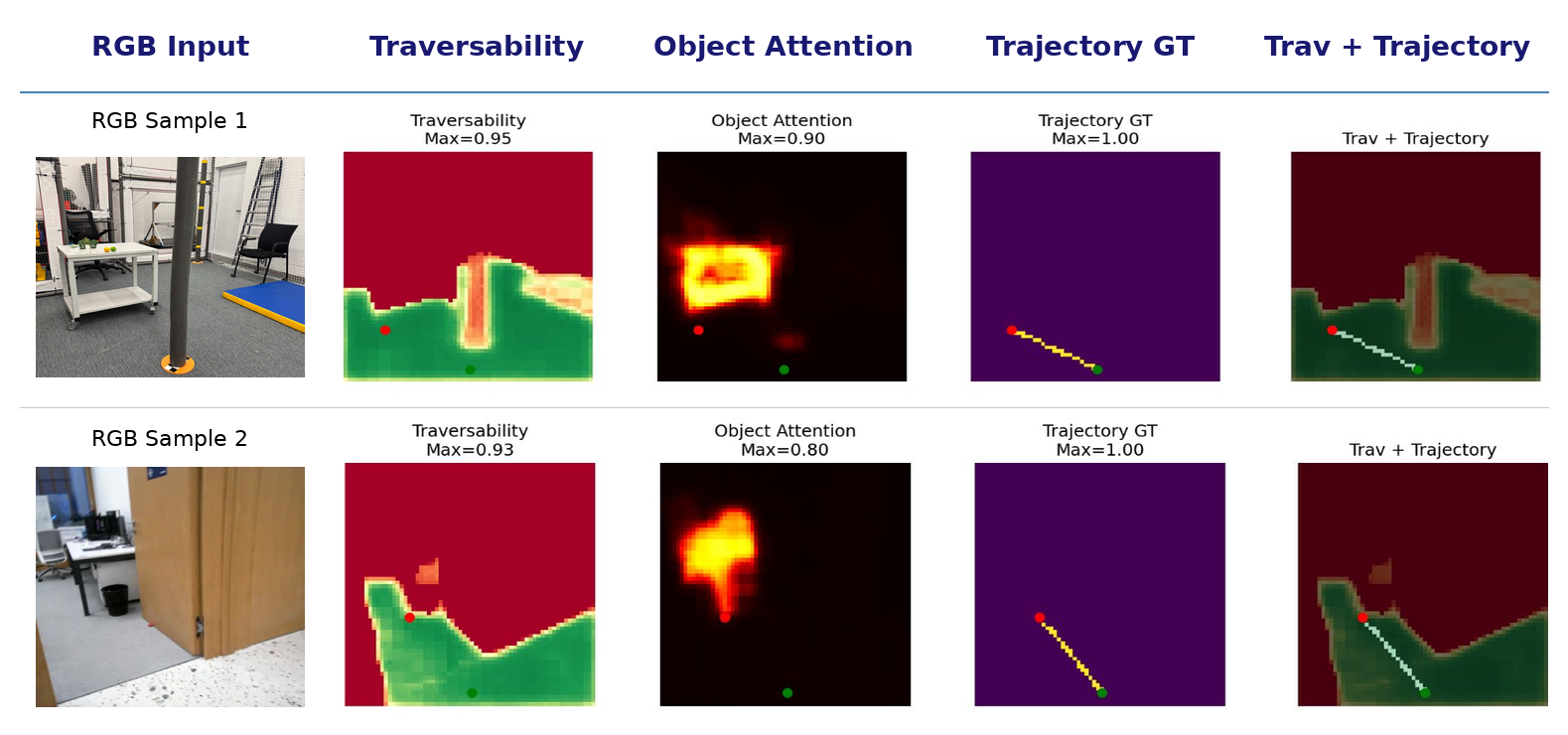}
    \caption{Ground truth generation pipeline for start and goal prediction. 
    For each sample: RGB input (64$\times$64), traversability map, 
    object attention heatmap, ground-truth trajectory, 
    and trajectory overlaid on the traversability map.}
    \label{fig:gt_pipeline}
\end{figure*}

Ground truth goals are generated hierarchically via AnyTraverse~\cite{sahu2025anytraverse} attention maps (Fig.~\ref{fig:gt_pipeline}). When task-relevant objects are detected, goals are placed toward them; otherwise, goals default to high-traversability forward regions. Start positions are sampled from traversable locations.

The ViT architecture enables dual modes: autonomous goal selection from scene context, or user-specified goals via feature matching between reference and scene images (batch size 2). This enables zero-shot navigation to novel objects without retraining or heavy VLMs at inference, maintaining lightweight deployment (2.0GB, 10 Hz).
\textbf{Auxiliary reconstruction losses.} To strengthen the model's understanding of scene geometry and semantic attention, we introduce auxiliary supervision on traversability maps and attention maps alongside the primary trajectory prediction objective. The model predicts traversability $\hat{\mathbf{T}}_t$ and attention $\hat{\mathbf{A}}_t$ as intermediate outputs from the depth-scale modulated decoder, supervised with:
\begin{equation}
\mathcal{L}_{\text{trav}} = \text{BCE}(\hat{\mathbf{T}}_t, \mathbf{T}_t^{\text{GT}}), \quad
\mathcal{L}_{\text{attn}} = \|\hat{\mathbf{A}}_t - \mathbf{A}_t^{\text{GT}}\|_2^2,
\end{equation}
where $\mathbf{T}_t^{\text{GT}}$ is the ground truth traversability from AnyTraverse and $\mathbf{A}_t^{\text{GT}}$ is the object attention map. These auxiliary losses encourage the encoder to learn robust geometric and semantic representations, which are then leveraged by the diffusion UNet through FiLM conditioning. The total training objective combines trajectory denoising, depth prediction, and auxiliary reconstruction:
\begin{equation}
\mathcal{L}_{\text{total}} = \mathcal{L}_{\text{diff}} + \lambda_{\text{depth}} \mathcal{L}_{\text{depth}} + \lambda_{\text{trav}} \mathcal{L}_{\text{trav}} + \lambda_{\text{attn}} \mathcal{L}_{\text{attn}},
\end{equation}
where $\lambda_{\text{depth}} = 0.1$, $\lambda_{\text{trav}} = 0.5$, and $\lambda_{\text{attn}} = 0.3$ balance the contribution of each supervision signal.
\FloatBarrier
This creates a natural hierarchy in goal generation. When a target object is present, the model operates in an \textit{object-directed} mode, placing the goal toward the detected object of interest. When no relevant object is identified, the model falls back to a \textit{traversability-directed} mode, placing the goal in the most traversable forward region. The model learns to internalize this priority during training, allowing it to reproduce the same hierarchy at inference without explicit scene classification.

\subsubsection{Depth-guided 3D reasoning}

To enable metric 3D reasoning while maintaining computational efficiency, we introduce depth-scale conditioning and trajectory-aligned depth prediction.

\textbf{Depth scale conditioning and supervision.} The model handles two metric regimes through depth scale conditioning $s \in \{\text{meter}, \text{cm}\}$. Ground truth depth sources are task-specific: for navigation, we distill meter-scale depth from ML-Depth-Pro~\cite{Bochkovskii2024} along trajectory waypoints for obstacle avoidance; for manipulation, we use known object ground truth poses to provide centimeter-scale depth for precise approach planning.

\textbf{Trajectory-aligned depth prediction.} We predict depth $\{\hat{d}_k\}_{k=1}^K$ only at trajectory waypoints $\{\mathbf{p}_k\}_{k=1}^K$ rather than across entire images, focusing supervision on task-relevant regions:
\begin{equation}
\mathcal{L}_{\text{depth}} = \frac{1}{K} \sum_{k=1}^K \|\hat{d}_k - d_k^{\text{GT}}\|_1,
\end{equation}
where $d_k^{\text{GT}}$ comes from supervision signal. This design reduces the computational burden while focusing metric understanding precisely where planning occurs. At inference, predicted depths enable collision checking without external sensors.

\section{Experiments and Results}
\subsection{Dataset Collection and Automated Supervision}

We collected RGB video at 30\,Hz from two scenes: pre-grasping scenes (humanoid robot, tabletop with diverse objects) and navigation scenes (humanoid robot, indoor corridors and towards different tables).

\textbf{Self-supervised supervision.} AnyTraverse~\cite{sahu2025anytraverse} generates traversability and attention maps using general prompts, enabling cross-task generalization. Ground truth trajectories are simulated using traversability guidance (collision avoidance) and attention guidance (goal direction), requiring no manual demonstrations.

Depth supervision adapts by task: navigation uses ML-Depth-Pro~\cite{Bochkovskii2024} for meter-scale monocular depth; for pre-grasping task, we use stereo depth prediction, providing precise spatial localization for approach trajectories.

\subsection{Training Strategy}

The model is trained jointly on three tasks (navigation, goal-directed navigation, pre-grasping) for 50 epochs using 20\,min video, batch size 32), taking 2 hours on a single RTX 4090 GPU and occupying memory of 6GB. 
Context-aware conditioning enables a single unified architecture to handle all operational modes without task-specific parameters.

\textbf{Rapid task adaptation.} To introduce a new task, we collect 5\,min of new visual data covering diverse scene variations with relevant objects of interest. Ground truth supervision (traversability, attention, trajectories) is generated automatically on-the-fly using AnyTraverse, while  manual annotation or teleoperation is not required. Finetuning the pretrained model for 10 epochs takes approximately 15-20\,min on RTX 4090, demonstrating effective cross-task knowledge transfer.

\subsection{Robot Platform}
We deploy our method on a Unitree G1 humanoid robot equipped with an RGB camera. The platform handles both manipulation (7-DOF arm) and navigation (bipedal locomotion) tasks. Inference runs on an NVIDIA GeForce RTX 4090 GPU, requiring only 2GB memory at 10\,Hz real-time performance, enabling efficient onboard autonomous operation.

\subsection{Baselines}
To evaluate the efficacy of our approach, we benchmark against state-of-the-art navigation and manipulation baselines. For navigation tasks (exploration and goal-directed), we compare against NoMaD~\cite{sridhar2023nomad}, a visual navigation policy trained on large-scale robotic datasets that generates subgoal waypoints from RGB observations. 

For manipulation and loco-manipulation scenarios, we initially attempted comparison with NVIDIA's GR00T architectures (n1.5 and n1.6)~\cite{gr00tn1_2025} deployed on a Unitree G1 humanoid embodiment.

\subsection{Real-World Deployment}
We evaluate three scenarios: (1) exploration navigation, (2) goal-directed navigation, and (3) pre-grasping planning.

\begin{figure}[H]
\centering
\includegraphics[width=0.9\columnwidth]{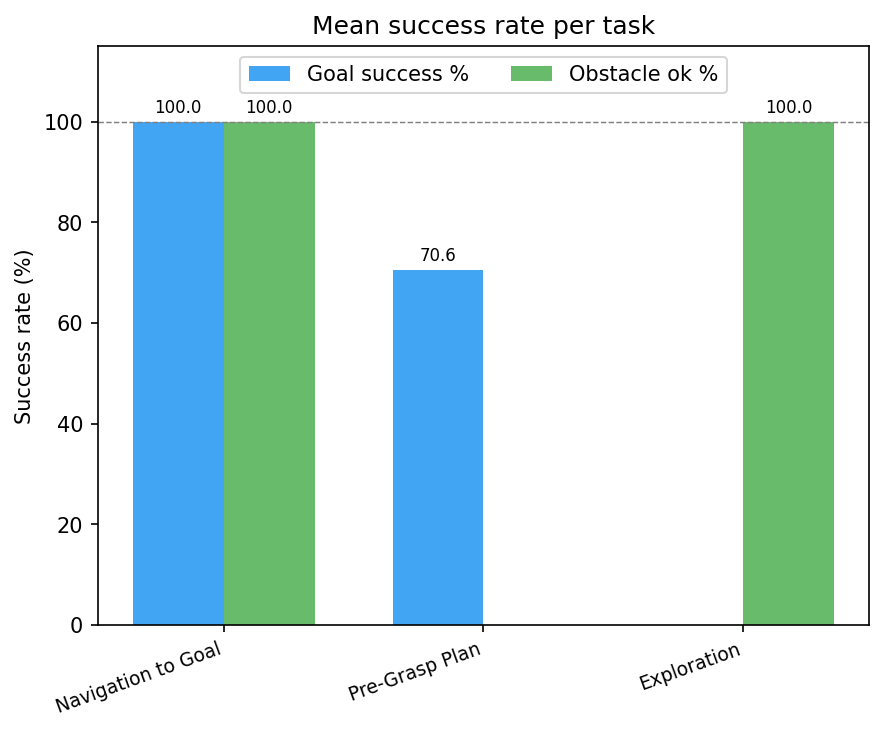}
\caption{Task performance on novel test scenes. Navigation: 100\% goal and collision success. Pre-grasping: 70.6\% goal success, 100\% collision-free. Exploration: 100\% obstacle avoidance. Zero-shot generalization across all modes.}
\label{fig:success_rates}
\end{figure}

A single checkpoint handles all modes without finetuning (Fig.~\ref{fig:success_rates}). Task switching occurs through context conditioning at inference: task mode, depth scale, and attention. This validates cross-task transfer while maintaining appropriate precision (meter vs. centimeter scale).

\subsection{Qualitative Results}

\begin{figure*}[h]
\centering
\includegraphics[width=0.5\textwidth]{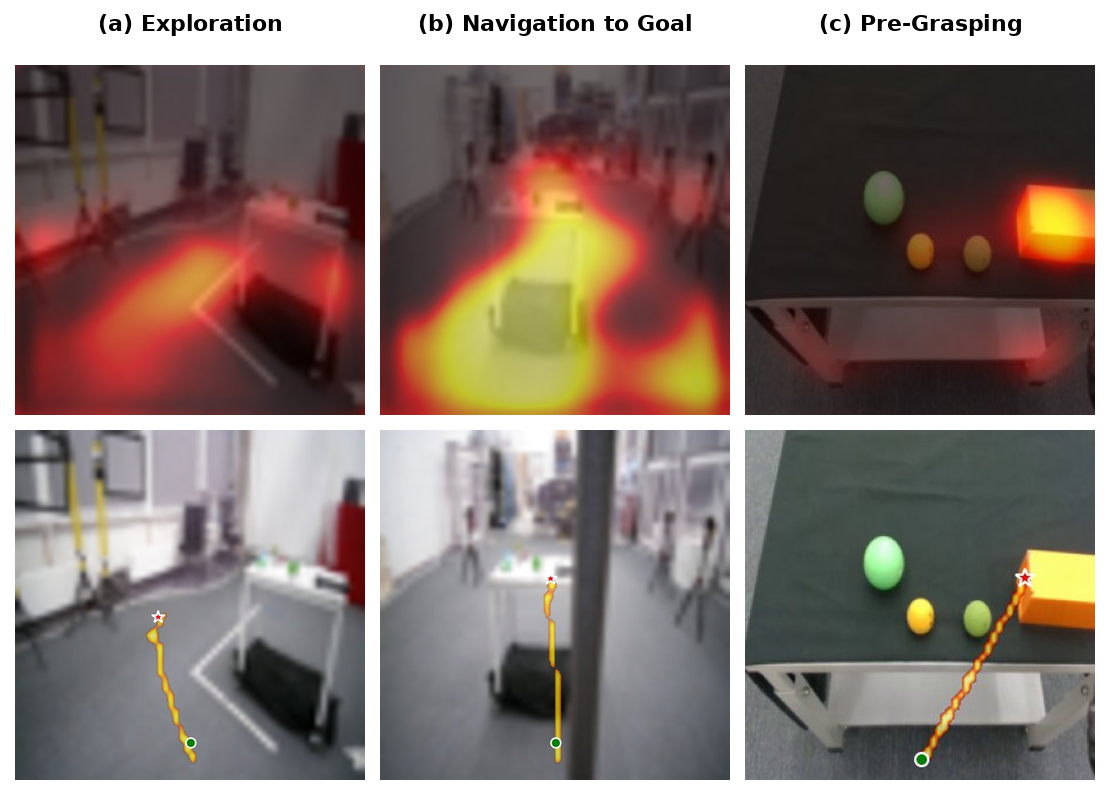}
\caption{Qualitative results showing attention maps (top) and trajectories (bottom) across three tasks. \textbf{(a) Exploration}: Floor attention identifies traversable regions for obstacle-free navigation. \textbf{(b) Navigation to Goal}: Attention highlights target table while generating goal-directed trajectory. \textbf{(c) Pre-Grasping}: Object-centric attention enables centimeter-precise approach planning. Green circles: start; red stars: predicted goals. All results demonstrate zero-shot generalization on novel test scenes.}
\label{fig:qualitative}
\end{figure*}

In Fig.~\ref{fig:qualitative} the task-adaptive attention across modes is demonstrated. \textbf{Exploration} (a): attention highlights traversable floor regions for obstacle avoidance. \textbf{Goal-directed navigation} (b): attention focuses on target tables while maintaining floor awareness. \textbf{Pre-grasping} (c): attention becomes object-centric for centimeter-precise approaches. The model autonomously predicts start positions (green), goals (red/orange stars), and collision-free trajectories from RGB only.

A single compact architecture unifies three distinct spatial attention representations: floor traversability for exploration, goal-region localization for navigation, and object-centric focus for manipulation. Despite this versatility, the model remains lightweight (2.0 GB, 10\,Hz inference) with rapid training (15-20\,min for new tasks). Context conditioning modulates attention patterns and spatial precision without architectural modifications.


\subsection{Experimental Results}

\begin{figure}[t]
\centering
\includegraphics[width=0.8\columnwidth]{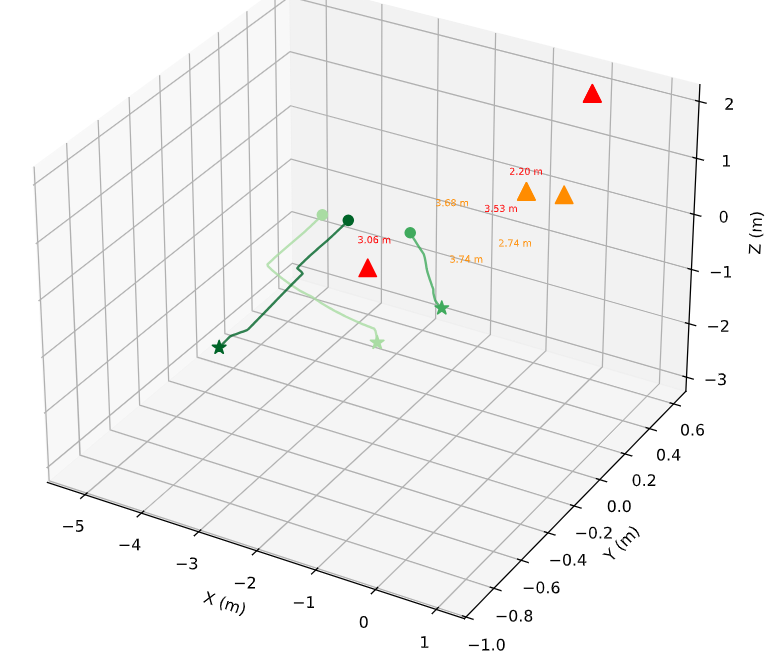}
\caption{3D trajectory prediction with obstacle avoidance. The model generates collision-free paths (green) maintaining mean 2.98\,mclearance from obstacles (minimum 2.20\,m). Vertical and lateral trajectory variations validate depth-based 3D spatial reasoning.}
\label{fig:trajectory_3d}
\end{figure}
\subsubsection{Exploration navigation}

Table~\ref{tab:navigation_results} compares navigation results against NoMaD~\cite{sridhar2023nomad}. Our model achieves 100\% obstacle avoidance during exploration, outperforming NoMaD (93\%) with larger safety margins (2.98\,m vs 2.24\,m). Learned floor attention identifies traversable regions from RGB only (Fig.~\ref{fig:trajectory_3d}), maintaining safe clearances (mean 2.98\,m, min 2.29\,m) and validating zero-shot generalization to unseen scenes.
\subsubsection{Goal-oriented navigation}

Our approach achieves 100\% autonomous goal-directed navigation with 0.29\,m accuracy (Table~\ref{tab:navigation_results}). NoMaD generates navigation waypoints but requires explicit goal images for object-directed navigation and cannot autonomously identify semantic targets (e.g., tables, chairs), precluding comparison on autonomous semantic goal selection.

When provided goal images, NoMaD achieves 60\% success at 0.5\,m, while our attention-guided approach achieves 52\% at 0.35\,m with 100\% collision-free navigation. Our unified model trained on 20\,min of cross-task data achieves competitive performance with NoMaD (trained on large-scale dataset) while offering exploration, autonomous semantic goal selection, and attention-guided navigation for novel objects in one architecture. The 0.35\,m approach supports manipulation interactions, with unsuccessful trials maintaining safety rather than forcing close approach, balancing interaction requirements with collision avoidance.

\begin{table}[h]
\centering
\caption{Navigation Performance. SR: Success Rate; Dist: mean clearance (exploration) / goal distance (navigation). $^*$N/A: NoMaD not trained for autonomous semantic goals. Goal success: $\leq$0.5\,m for interaction.}
\small
\begin{tabular}{lccc}
\toprule
\textbf{Scenario} & \textbf{Method} & \textbf{SR (\%)} & \textbf{Dist (m)} \\
\midrule
\multirow{2}{*}{Exploration} 
& NoMaD & 93 & 2.24 \\
& Ours & \cellcolor{green!20}\textbf{100} & \cellcolor{green!20}\textbf{2.98} \\
\midrule
\multirow{2}{*}{Autonomous Goal} 
& NoMaD & N/A$^*$ & N/A \\
& Ours & \cellcolor{green!20}\textbf{100} & \cellcolor{green!20}\textbf{0.29} \\
\midrule
\multirow{2}{*}{Provide goal} 
& NoMaD & 60 & 0.5 \\
& Ours & 52 & \cellcolor{green!20}\textbf{0.35} \\
\bottomrule
\end{tabular}

\label{tab:navigation_results}
\vspace*{-2mm}
\end{table}

\subsubsection{Pre-grasping motion planning}

\begin{figure}[h]
\centering
\includegraphics[width=0.9\columnwidth]{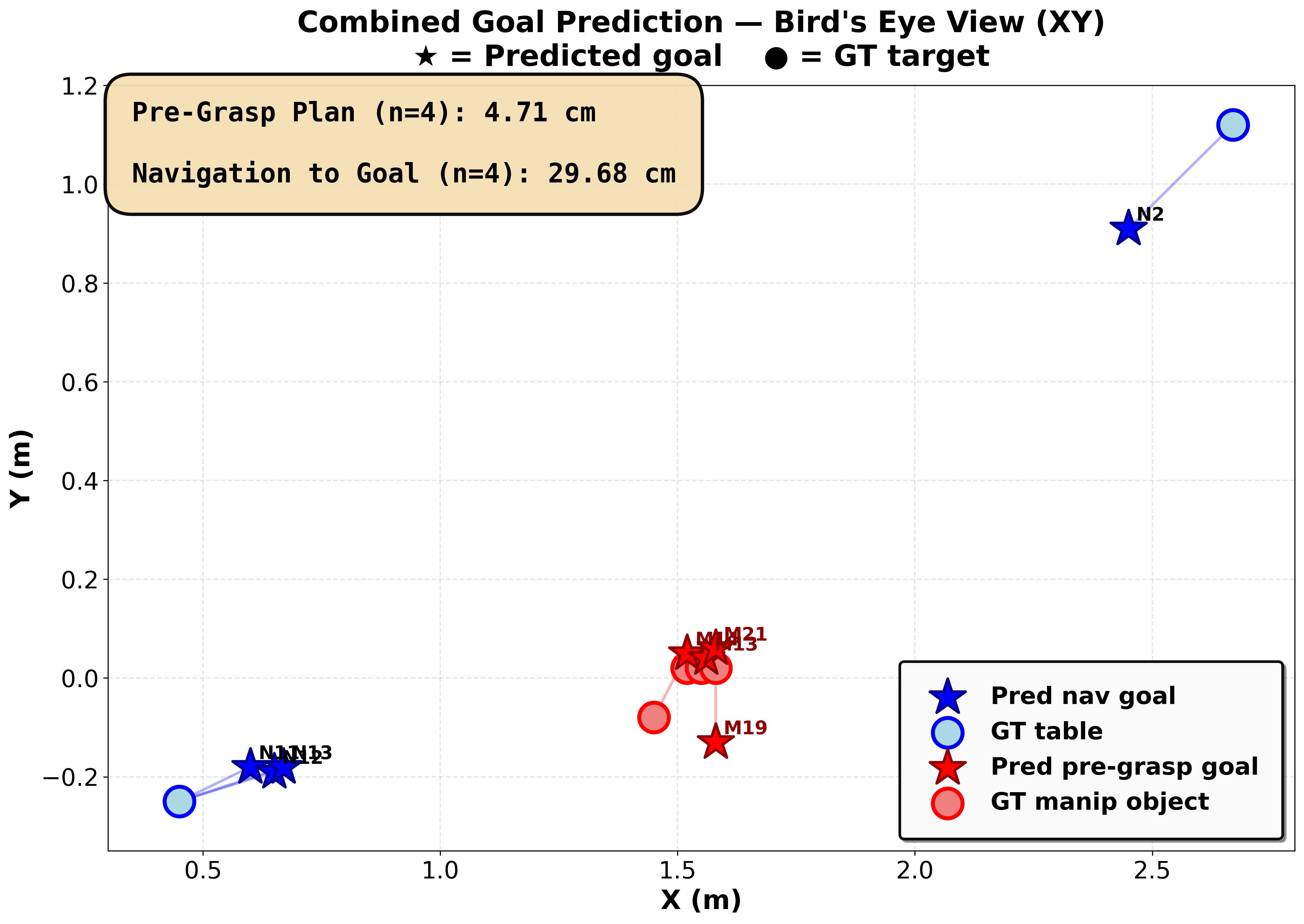}
\caption{Goal prediction accuracy (bird's-eye view, $n=8$). Pre-grasping achieves 4.71\,cm mean error versus 29.68\,cm for navigation, demonstrating task-appropriate precision through depth-scale conditioning: centimeter-level for manipulation, decimeter-level for navigation.}
\label{fig:goal_prediction}
\end{figure}

Pre-grasping achieves 70.6\% goal success with 100\% collision-free execution and 4.71\,cm mean error. This demonstrates centimeter-scale precision compared to meter-scale navigation (29.68\,cm)—a 6× improvement validating depth-scale conditioning (Fig. \ref{fig:goal_prediction}). Object-centric attention enables precise approach trajectories from RGB only, with perfect collision avoidance despite localization challenges. The unified model's ability to switch between operational scales through context alone demonstrates effective cross-task knowledge transfer.

\subsection{Comparison with Vision-Language-Action Baselines}

\textbf{Stationary manipulation (GR00T n1.5):} For the baseline manipulation task, the robot was commanded to ``pick up an apple and place it in a blue plate". The GR00T n1.5 baseline model was fine-tuned over an extensive dataset of 1,100 expert demonstrations (averaging 15 seconds per episode, totaling approximately 4.5 hours of continuous teleoperation data). Under this regime, the baseline achieved a 60\% success rate in simulation. However, during real-world zero-shot deployments, performance degraded to a 40\% success rate (4/10 successful trials). This drop in sim-to-real transfer highlights that while GR00T provides a strong semantic prior, its end-to-end nature remains highly susceptible to grasp instability under slight physical pose variations.

\textbf{Zero-shot generalization failure in GR00T.} We attempted to compare against NVIDIA's GR00T n1.6~\cite{gr00tn1_2025} for mobile manipulation tasks. However, GR00T exhibited critical limitations: (1) only 27-32\% success rate on our loco-manipulation benchmark (vs. 58\% reported on official tasks), and (2) **complete failure to generalize zero-shot to novel scene configurations** outside its training distribution. This prevented meaningful comparison across our diverse test scenarios featuring unseen object arrangements and spatial layouts. The failure mode is systematic: without explicit geometric reasoning modules (traversability, spatial attention), end-to-end VLA models memorize training scene patterns rather than learning transferable spatial understanding, causing catastrophic performance degradation under distribution shift.

\textbf{Explicit geometric reasoning enables generalization.} Our approach addresses this limitation through structured geometric reasoning: traversability maps for safe navigation and spatial attention for goal-directed planning. This design enables zero-shot transfer to novel scenes from minimal training data (5\,min of auto-supervised images per task), achieving robust performance where end-to-end VLAs fail.


\section{Conclusion and Future Work}

We presented a unified image-space diffusion policy bridging meter-scale navigation and centimeter-scale pre-grasping. A single model achieves 10\,Hz performance with task-appropriate precision (4.71\,cm) through multi-scale FiLM conditioning and trajectory-aligned depth reasoning, demonstrating robust zero-shot generalization versus compute-heavy VLA models.

The ViT-based attention predictor enables autonomous goal selection and user-specified navigation via feature matching, supporting zero-shot behavior toward novel objects.

Limitations include AnyTraverse supervision dependency and lack of force feedback. Future online learning will enhance AnyTraverse with self-supervised robot experience: the strong visual encoder enables feature similarity learning, while traversability and pre-grasping accuracy refine through execution feedback (collisions, grasp outcomes). The robot deploys with strong priors from offline training, then efficiently adapts through interaction. Additional directions include multimodal sensing and bimanual coordination. Validating incremental learning and lifelong semantic expansion remain key for deployment.

\bibliographystyle{ieeetr}
\bibliography{references}

\end{document}